\documentclass[letterpaper, 10 pt, conference]{ieeeconf}

\IEEEoverridecommandlockouts                              

\usepackage{xcolor}
\usepackage{graphics} 
\usepackage{epsfig} 
\usepackage{mathptmx} 
\usepackage{times} 
\usepackage{amsmath} 
\usepackage{amssymb}  
\usepackage{subcaption}
\usepackage[switch]{lineno}
\usepackage{multirow}
\usepackage{booktabs}
\usepackage{acro}
\usepackage{amssymb}
\usepackage{bm}

\DeclareAcronym{UAV}{
short=UAV,
long=unmanned aerial vehicles
}

\DeclareAcronym{RA}{
short=RA,
long=Residual Aligner
}

\DeclareAcronym{MA}{
short=MA,
long=Motion Aware
}

\DeclareAcronym{SFM}{
short=SfM,
long=Structure from Motion
}

\DeclareAcronym{CNN}{
short=CNN,
long= Convolutional Neural Networks
}

\DeclareAcronym{GT}{
short=GT,
long=Ground Truth
}

\DeclareAcronym{F2D}{
short=F2D,
long=Flow to Depth
}

\DeclareAcronym{D2F}{
short=D2F,
long=Depth to Flow
}

\DeclareAcronym{FPS}{
short=FPS,
long=Frames Per Second
}

\DeclareAcronym{EMA}{
short=EMA,
long=Exponential Moving Average
}

\title{\LARGE \bf
A Compacted Structure for Cross-domain learning on Monocular Depth and Flow Estimation
}

\author{Yu Chen$^{\ast 1}$, Xu Cao$^{\ast 2}$, Xiaoyi Lin$^{3}$, Baoru Huang$^{4}$, Xiao-Yun Zhou$^{5}$, Jian-Qing Zheng$^{6}$, Guang-Zhong Yang$^{7}$ 
\thanks{$\ast$Yu Chen and Xu Cao contribute equally to this paper}%
\thanks{Corresponding author: Jian-Qing Zheng.}
\thanks{$^{1}$Yu Chen is with Robotics Institute, Carnegie Mellon University, Pittsburgh, USA, 15213
        {\tt\small yuchen3@CS.CMU.EDU}}%
\thanks{$^{2}$Xu Cao is with The Hong Kong University of Science and Technology (Guangzhou), No. 1, Duxue Road, Nansha District, Guangzhou  511458, China
        {\tt\small xcao635@connect.hkust-gz.edu.cn}}%
\thanks{$^{3}$Xiaoyi Lin is with Tandon School of Engineering, New York University Brooklyn, NY, USA, 11201
        {\tt\small linxiaoyi1108@gmail.com}}%
\thanks{$^{4}$Baoru Huang is with Department of Surgery and Cancer, Imperial College London,           London, UK, SW7 2AZ
        {\tt\small baoru.huang18@imperial.ac.uk}}%
\thanks{$^{5}$Xiao-Yun Zhou is with Amazon
        {\tt\small xiaoyun.zhou27@gmail.com}}%
\thanks{$^{6}$Jian-Qing Zheng is with the Kennedy Institute of Rheumatology and the Big Data Institute, Oxford, UK
        {\tt\small jianqing.zheng@some.ox.ac.uk}}%
\thanks{$^{7}$Guang-Zhong Yang is with the Institute of Medical Robotics, Shanghai Jiao Tong University, No. 800, Dongchuan Road, Minhang District, Shanghai 200240, China
        {\tt\small  gzyang@sjtu.edu.cn}}
}

\begin{document}

\maketitle
\thispagestyle{empty}
\pagestyle{empty}

\begin{abstract}

Accurate motion and depth recovery is important for many robot vision tasks including autonomous driving. Most previous studies have achieved cooperative multi-task interaction via either pre-defined loss functions or cross-domain prediction. This paper presents a multi-task scheme that achieves mutual assistance by means of our \ac{F2D}, \ac{D2F}, and \ac{EMA}. \ac{F2D} and \ac{D2F} mechanisms enable multi-scale information integration between optical flow and depth domain based on differentiable shallow nets. A dual-head mechanism is used to predict optical flow for rigid and non-rigid motion based on a divide-and-conquer manner, which significantly improves the optical flow estimation performance. Furthermore, to make the prediction more robust and stable, \ac{EMA} is used for our multi-task training. Experimental results on KITTI datasets show that our multi-task scheme outperforms other multi-task schemes and provide marked improvements on the prediction results.

\end{abstract}


\section{INTRODUCTION}

\begin{figure}[htpb]
    \centering
    \includegraphics[width=1\linewidth]{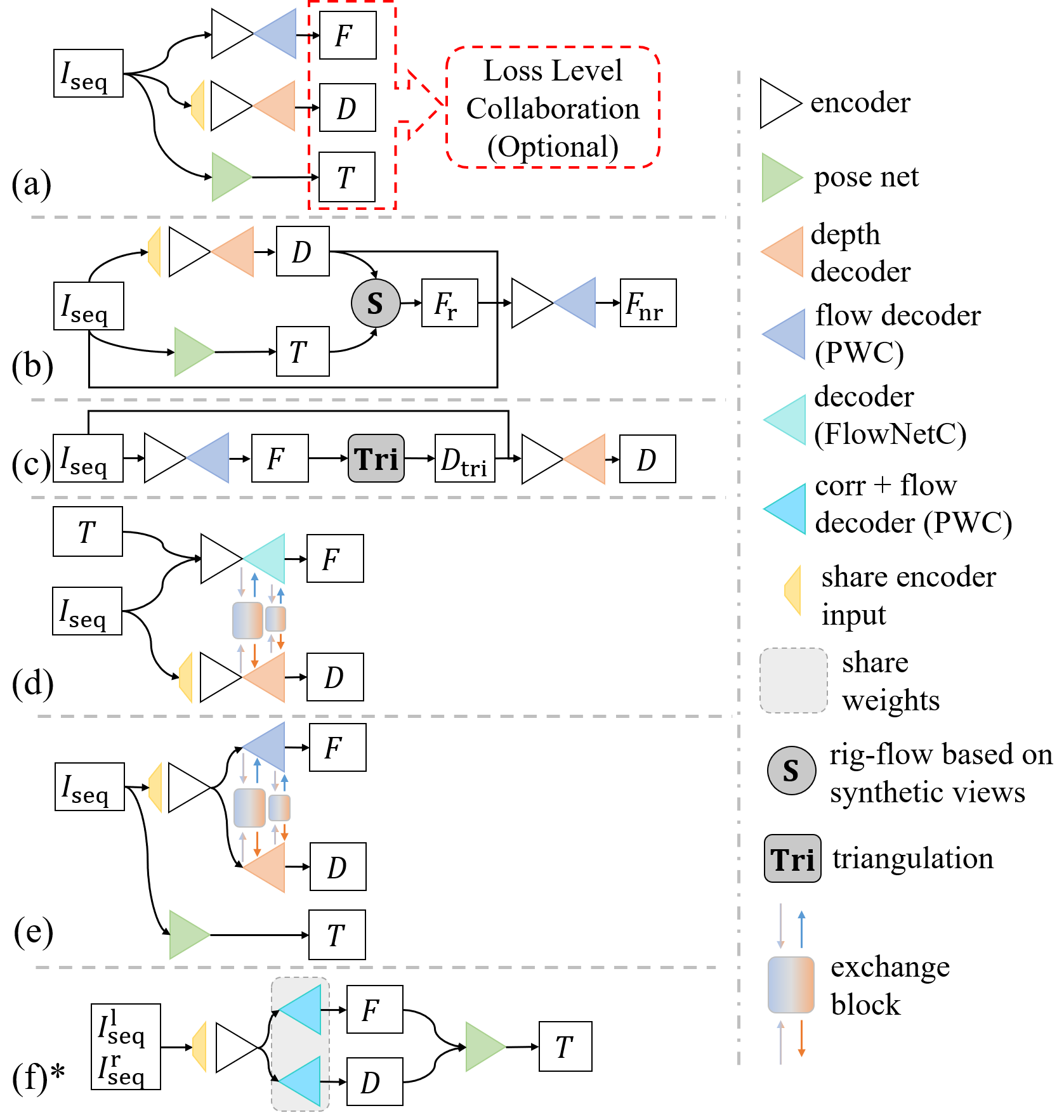}
    \caption{Unsupervised multi-task scheme. (a) Task specified separate network scheme supervised only by 3D geometry based loss functions (share encoder input: the encoder processes each images separately and the same below); (b) Flow estimation based on the rigid flow from synthetic views and original image sequence; (c) Triangulated depth based on estimated optical flow; (d) and (e) Information exchange block based coupled depth and optical flow network with/without known pose; (f) Joint stereo depth and optical flow estimation using shared weight decoder, combined with pose estimation based on the above results. (* refers to stereo depth different from monocular depth in other schemes)}
    \label{multitask_brief_fig}
\end{figure}

3D scene understanding from single view video sequences is an important topic in robot vision, which consists of several classic computer vision tasks, including depth estimation, optical flow estimation, and visual odometry prediction. It plays a key role in many real-world applications, including autonomous driving \cite{janai2020computer}, \ac{UAV} \cite{pirvu2021depth}, and surgical navigation \cite{huang2021self,zheng2019real}. 

\textbf{Depth Estimation.} Traditional method referred to as \ac{SFM} addresses depth and camera pose estimation simultaneously by keypoint matching. However, traditional methods tend to use hand-crafted features and often rely on sparse correspondence of low-level features \cite{klodt2018supervising,Yang_2018_ECCV}. 

More recently, \ac{CNN} have shown to be able to establish dense mapping relationships between single view images and the corresponding depth information in a supervised manner \cite{eigen2014depth,laina2016deeper,li2018monocular}. Nevertheless, the requirement of depth annotations is a major challenge for supervised methods \cite{2014Depth,2008Make3D, huang2022self}. Without any dependence on depth annotations, self-supervised methods \cite{garg2016unsupervised,godard2017unsupervised, huang2022h} are trained with only stereo images by leveraging the multi-view geometric constraints and capable of monocular depth estimation. Zhou \textit{et al.} \cite{zhou2017unsupervised} proposed a self-supervised monocular depth estimation paradigm by combining ego pose estimation and depth estimation tasks. By leveraging the appearance similarity indicators calculated by synthetic views \cite{zhou2016view} from temporal continuous frames and the current frame, these methods show promising performance \textit{w.r.t.} other self-supervised methods. However, simple appearance loss is insufficient as supervision especially in illumination variations scenes. To this end, Xiong et al. \cite{xiong2021self} proposed several novel scale-consistent geometric constraints by jointly considering relative forward-backward poses consistency and depth reconstruction error to further improve the depth estimation performance.

\textbf{Optical Flow Estimation}. Optical flow estimation was traditionally be treated as an optimization problem over the dense displacement fields between temporal adjacent image pairs \cite{horn1981determining,chen2016full}. The optimization objective consists of two main terms: 1) \textit{data}  -- that encourages correlations in visually similar regions; 2) \textit{regularization} -- that penalizes incoherent motion fields according to the motion priors. Whilst offering good performance in general, the inherent optimization process entails high computational cost making these methods difficult to be used for real-time applications \cite{sun2018pwc}. 


Recently, \ac{CNN}-based approaches have shown improved optical flow estimation results \cite{ilg2017flownet,sun2018pwc,teed2020raft}. However, these supervised methods require dense per-pixel optical flow annotation, which is laborious. By optimizing photometric consistency and local flow smoothness, unsupervised methods overcome the need for annotations. Several recent methods have greatly improved prediction results by checking forward-backward consistency \cite{sundaram2010dense}, filtering range map to omit occlusion regions \cite{wang2018occlusion} or using loss function based on occlusion-aware bidirectional flow estimation and the robust census transform \cite{meister2018unflow}. Unlike coarse-to-fine optical flow prediction, such as PWC-Net \cite{sun2018pwc}, more accurate results have been obtained by methods applying RNN structures to optical flow prediction scenarios \cite{teed2020raft}.

\textbf{Unsupervised Multi-task Network Scheme}. Depth and optical flow can be estimated in a joint manner due to their inherent geometric correspondences. Many approaches \cite{zou2018df, chen2019self, wang2019unos} have predicted depth and optical flow with separate networks trained by geometric constraints based consistency losses. Other methods \cite{ranjan2019competitive,wang2020unsupervised,cao2021robust,zhao2020towards} have combined depth and optical flow networks by serving depth or optical flow estimation results as the 3D information supervision of the other task. Although efficient, the above methods omit the inherent correspondence between latent information of different domains, which also serve as important assistance to multi-task estimation.

To further improve the multi-task network performance, we propose a novel multi-task scheme for depth estimation and optical flow prediction. It enables a bidirectional interaction of latent information between the depth and optical flow networks, enabling mutual assistance between two tasks. Our contributions can
be summarised as follows:

\begin{itemize}
    \item We proposed \textbf{\ac{D2F}} and \textbf{\ac{F2D}} mechanisms, which allow the latent information of depth and optical flow prediction transfer to and assist the estimation of the other task.
    \item For the former mechanism, we adopt the \ac{EMA} to the multi-task learning and further improve the performance of cross-task collaboration.
    \item Dual-head mechanism is designed to enhance the optical flow prediction results by implicitly disentangling the representation of foreground and background.
\end{itemize}


To prove the effectiveness of our multi-task scheme, experiments are conducted on \textbf{KITTI} \cite{geiger2012we} datasets. The experiments cover different combinations of \ac{D2F}, \ac{F2D}, \ac{EMA} training approach, and dual-head mechanism based on the baseline scheme in Fig.~\ref{multitask_brief_fig}(a). \ac{F2D} and \ac{D2F} enable the interchange of latent information between depth and optical flow estimation. Experiment results show that our \ac{D2F} mechanism can significantly improve the optical flow prediction accuracy. At the same time, the road signs and streetlamps are more clearly told in depth estimation. Experiments also show that our \ac{F2D} can further improve depth estimation accuracy. We further show that our \ac{EMA} training approach can improve both depth and optical flow prediction results. Our dual-head mechanism enables the division of the motion field into rigid and non-rigid parts based on a \textbf{divide-and-conquer} manner and improves the optical flow accuracy significantly.

\begin{figure*}[htpb]
    \centering
    \includegraphics[width=1\linewidth]{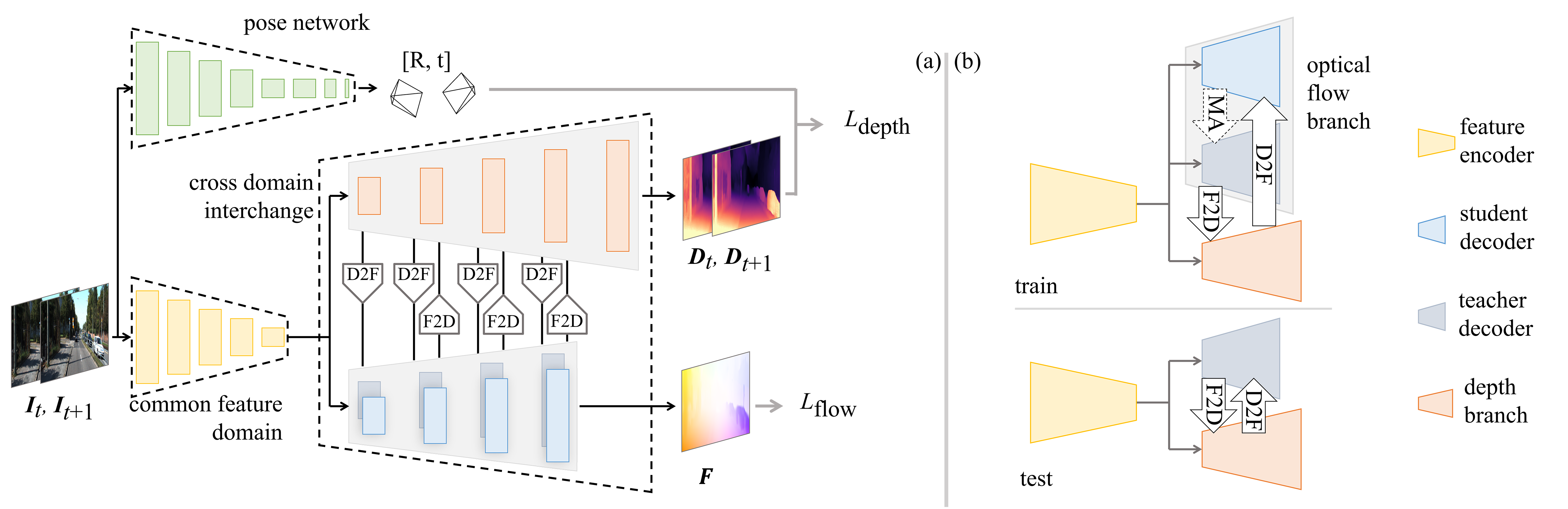}
    \caption{The overall architecture of the proposed network.}
    \label{fig:network}
\end{figure*}

\section{RELATED WORK}

\subsection{Loss-level Collaboration in Parallel Multi-task Network}

Previous methods achieve cross-task interaction by applying cross-domain loss functions to multi-task learning scheme combined with separate single-task networks, as illustrated in Fig.~\ref{multitask_brief_fig}(a). Ranjan et al. \cite{ranjan2019competitive} introduced a competitive collaboration strategy that can classify moving objects and static background and effectively coordinate the training of the multi-task network. Benefiting from the framework, these tasks with inherent correspondence could reinforce each other. To reduce the violation of non-rigid motion flow caused by moving objects and occluded pixels, Wang et al. \cite{wang2020unsupervised} constructed various masks for a valid consistency loss calculation, which remarkably improves the estimation of monocular depth, ego pose, and optical flow in temporal consecutive frames. Cao et al. \cite{cao2021robust} proposed photometric loss accompanied with bundle adjustment modules to achieve more accurate results. Zhao et al. \cite{zhao2020towards} facilitated pose estimation by directly calculating camera pose from optical flow to improve depth estimation and optical flow prediction.

\subsection{Cross-task Boosting in Multi-task Network}

Other than cross-task collaboration based on the loss functions, other methods took further steps to boost multi-task performance by introducing information interaction between the depth and optical flow estimation network. GeoNet et al. \cite{yin2018geonet,yang2021unsupervised} derives rigid flow from depth and pose prediction, which is then fine-tuned by optical flow network as illustrated in Fig.~\ref{multitask_brief_fig}(b). By grafting the depth estimation network behind the flow-motion network, the method proposed by \cite{xie2020video} refines the camera poses from GPS, IMU, or odometry algorithms and depth estimation with the confidential map as illustrated in Fig.~\ref{multitask_brief_fig}(c). Yang et al. \cite{yang2021dense} have been estimated depth map based on optical flow from pretrained FlowNet with mid-point triangulation method. DRAFT \cite{guizilini2022learning} generates coarse depth prediction by triangulation from optical flow estimated by RAFT \cite{jia2021braft}. DRAFT then estimates the fine-grained depth and scene flow by joint leveraging coarse depth prediction and pyramid correlation. Different from the methods mentioned above, our cross-task exchange mechanism \ac{D2F} and \ac{F2D} enables multi-scale latent information interchange and facilitates multi-level cross-domain collaboration between depth and optical flow estimation. 

As for multi-task feature-level collaboration, Chi et al. \cite{chi2021feature} introduced a feature-level collaboration mechanism for the stereo depth estimation, flow prediction, and pose detection networks as illustrated in Fig.~\ref{multitask_brief_fig}(f). Whereas our scheme is dedicated to more challenging monocular depth estimation tasks, rather than stereo depth estimation. Hur et al. \cite{hur2020self} proposed an architecture to estimate depth and 3D motion simultaneously from a single decoder. In addition to the previous cross-task boosting scheme, DENAO \cite{chen2020denao} proposed multi-scale exchange blocks which deeply coupled depth and optical flow branches according to epipolar geometry constraints, as illustrated in Fig.~\ref{multitask_brief_fig}(d). The auxiliary optical flow improves the depth predictions and in turn yields large improvements in optical flow accuracy. However, by leveraging known ego camera pose and the proposed epipolar layer, DENAO is a partially supervised multi-task scheme. We proposed a fully unsupervised multi-task scheme with cross-task information exchange blocks. The geometry constraints between optical flow and depth predictions are leveraged implicitly, rather than the explicit 3D-to-2D projection and SVD decomposition-based triangulation in DENAO.

\section{Methodology}

Given a pair of consecutive frames $(\bm{I}_{t}, \bm{I}_{t+1})$ from an unlabeled monocular video, our method estimates optical flow $\bm{F}$ between the two frames $(\bm{I}_{t}, \bm{I}_{t+1})$ and the corresponding depth maps $\bm{D}_{t}$ and $\bm{D}_{t+1}$.

The overall architecture of the proposed method is illustrated in Fig.~\ref{fig:network}(a). The multi-task estimation network consists of a shared encoder extracting the common feature domain, a two-branch decoder regress in the depth and optical domain, and exchange blocks for cross-domain message interchanging. As described in Sec.~\ref{sec:cross_domain}, the two branches in the cross domain interchange, respectively for estimation of depth and optical flow, collaborate in their tasks by transferring cross-domain latent information from multiple layers to each other. 

\ac{EMA} is embedded into our training framework to improve the depth and optical flow estimation as shown in Fig.~\ref{fig:network}(b), with the detail described in Sec.~\ref{sec:mean_techer}. Following mean teacher method \cite{tarvainen2017mean}, the optical flow branch is further divided into student and teacher decoder. The structure of the two optical flow decoders is adopted from PWC-Net \cite{sun2018pwc} with dual-head mechanism (Sec.~\ref{sec:dual-head}) incorporated in each decoder layer. The depth branch contains a depth decoder following the decoder structure of monodepth \cite{godard2017monodepth}.

A separate camera pose prediction network is applied during training for jointly learning depth and camera pose based on geometric consistency.

\subsection{Cross-domain Collaboration}
\label{sec:cross_domain}

\begin{figure*}[htpb]
    \centering
    \includegraphics[width=1\linewidth]{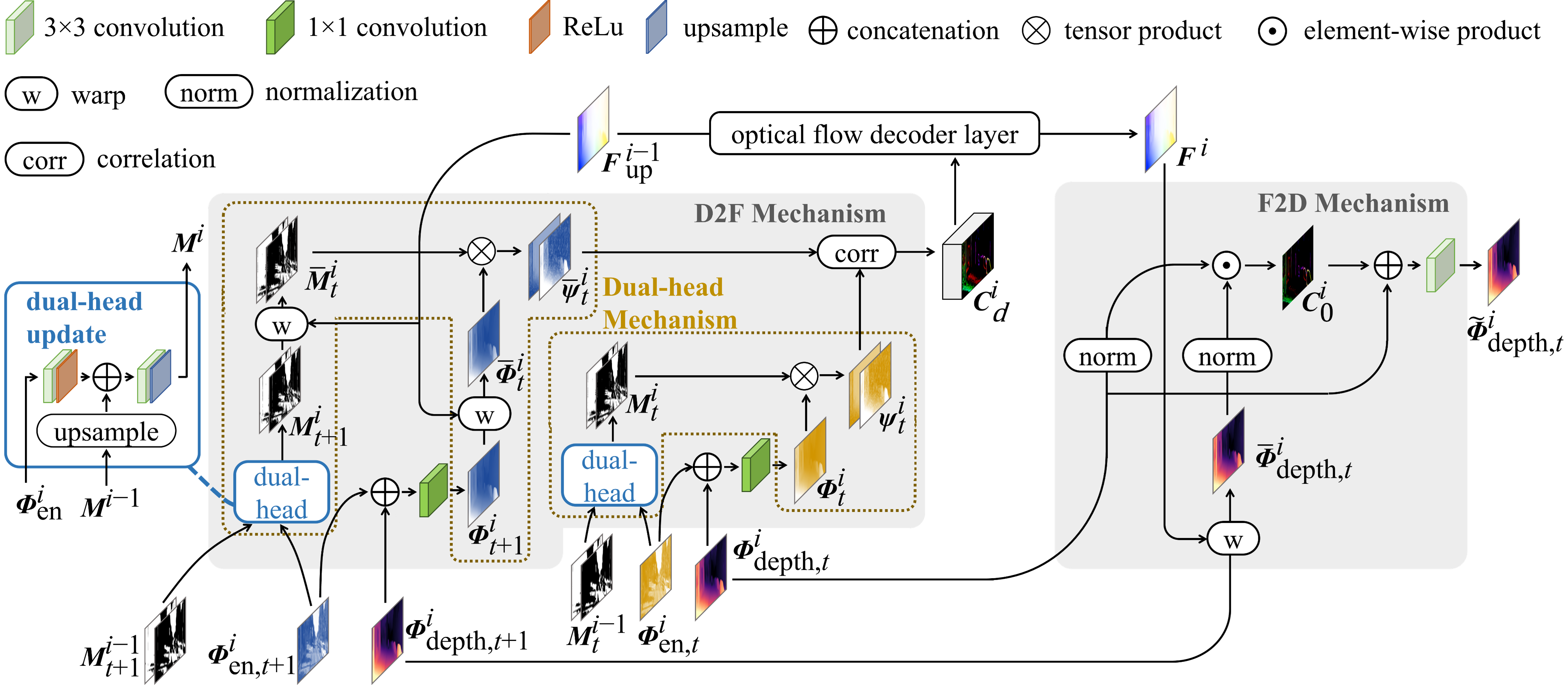}
    \caption{The architecture of $i^{\rm th}$ \ac{D2F}, \ac{F2D}, and Dual-head module. The \ac{D2F} module calculates cost volume $ \bm{C}$ from depth feature map $\bm{\Phi}_{\mathrm{depth}}^{i}$, encoder feature map $\bm{\Phi}_{\mathrm{en}}^{i}$, and dual-head mask $\bm{M}^{i-1}$. The \ac{F2D} mechanism refine depth feature map $\bm{\Phi}_{\mathrm{depth}}^{i}$ into $\tilde{\bm{\Phi}}_{\mathrm{depth}}^{i}$. The Dual-head module transfer fused feature map $\bm{\Phi}^{i}$ into dual-head feature map $\bm{\Psi}^{i}$.}
    \label{fig:mechanism}
\end{figure*}

Estimation of depth and optical flow both requires comprehensive understanding of stereo vision. We hypothesize that these two tasks can be solved from a common feature domain and thus utilize only one shared encoder.
For further mutually benefiting the two tasks in the exchange blocks, depth and optical flow branches collaborate with each other via \ac{D2F} and \ac{F2D} mechanisms. \ac{D2F} and \ac{F2D} mechanisms help to refine the prediction from multiple levels of the decoder layers by providing each other with cross-domain information, rather than merely supervising the final prediction of the entire network.

\subsubsection{\ac{D2F} mechanism}
\label{sec:d2f}


In solitary optical flow prediction, the network determines the pixel shift by comparing the RGB information of two images. Benefiting from \ac{D2F} mechanism, the depth branch provides additional geometric information for optical flow prediction.



\ac{D2F} mechanism provides depth information of the images for optical flow prediction. As shown in Fig.~\ref{fig:mechanism}, the $i^{\rm th}$ \ac{D2F} mechanism takes dual-head masks $\bm{M} ^{i-1}$ received from the $(i-1)^{\rm th}$ decoder layer, encoder feature maps $\bm{\Phi}_{en}^{i}$, and depth feature maps $\bm{\Phi}_{depth}^{i}$ from two frames of images as input to generate cost volume $\bm{C}_{d}^{i}$. Together with the cost volume, the up-sampled optical flow from the $(i-1)^{\rm th}$ decoder layer $\bm{F}^{i-1}$ is further refined by the $i^{\rm th}$ optical flow decoder layer to generate $\bm{F}^{i}$.


As the pre-process, the \ac{D2F} mechanism first generates fused feature map, $\bm{\Phi}^{i}$ by $\bm{\Phi}^{i} = \mathrm{\mathcal{C}^{1}}\left( \bm{\Phi}_{\rm en}^{i} \oplus \bm{\Phi}_{\rm depth}^{i} \right)$ where $\oplus$ denotes the channel concatenation and $\mathrm{\mathcal{C}^{1}}$ denotes 1$\times$1 convolutional layers. $\bm{\Phi}^{i}$ will be then operated by dual-head mechanism to produce refined feature map $\bm{\Psi}^{i}$, which is discussed in section \ref{sec:dual-head}. 


The cost volume $\bm{C}_{d}^{i}$ is obtained through the correlation between $\bm{\Psi}_{t}^{i}$ and $\overline{\bm{\Psi}}_{t}^{i}$ with a limited
range of $d$ pixels:
\begin{equation}
    \bm{C}_{d}^{i} \left( \bm{\Psi}_{t}^{i},\overline{\bm{\Psi}}_{t}^{i} \right) = \frac{1}{N} \left( \bm{\Psi}_{t}^{i} + \bm{o} \right)^\top \left(\overline{\bm{\Psi}}_{t}^{i} + \bm{o} \right)
    , \bm{o} \in [-d,d] \times [-d,d]
    \label{CV}
\end{equation}
where N denotes the length of the column vector $\bm{\Psi}_{t}^{i}$.

\subsubsection{\ac{F2D} Mechanism}
\label{sec:f2d}


The \ac{F2D} mechanism improves depth prediction with the stereo features extracted from optical flow.


As illustrated in Fig.~\ref{fig:mechanism}, the \ac{F2D} mechanism obtain $\bm{F}^{i}$ to help improve the depth feature map $\bm{\Phi}_{\rm depth}^{i}$ and produce the refined depth feature map $\tilde{\bm{\Phi}}_{\rm depth}^{i}$.



We first normalize the value of $\overline{\bm{\Phi}}^{i}$ to [0,1]. The normalized feature maps are then applied to calculate the cost map $\bm{C}_{0}$ following Equation~\eqref{CV} where $d$ is set to 0.


The output refined depth feature map $\tilde{\bm{\Phi}}_{t}^{i}$ is calculated with $\tilde{\bm{\Phi}}_{t}^{i} = \mathrm{\mathcal{C}^{3}} \left( \bm{\Phi}_{t}^{i} \oplus \bm{C}_{0} \right)$ where $\mathrm{\mathcal{C}^{3}}$ denotes the 3$\times$3 convolutional layer. The refined depth feature map will be obtained by the next depth decoder layer.


\subsection{\ac{EMA} for Cross-task Prediction}
\label{sec:mean_techer}

In our model, cross-domain information walks double-sided between depth and optical flow branches. Parameter fluctuation of any of these two branches will directly affect the other one. Moreover, \ac{D2F} and \ac{F2D} work at multiple decoder layers. The error of one layer will accumulate at the next layer, and will eventually cause significant bias on the final prediction result. These factors might cause the network not to converge easily to the optimal solution during training.

To make the prediction more robust and stable, we train and test the optical flow branch following \ac{EMA} approach. \ac{EMA} \cite{tarvainen2017mean} was adapted to the \ac{CNN} training to facilitate the semi-supervised classification result by averaging model weights instead of predictions. During training, the teacher model updates its parameters by using the \ac{EMA} weights of the student model. To the best of our knowledge, this is the first application of \ac{EMA} to cross-task prediction of depth and optical flow. 

As illustrated in Fig.~\ref{fig:network}(b), the optical flow branch contains teacher and student decoders. Both decoders share the same architecture and initial parameters as well. The teacher decoder only works during training. During training, the student branch updates its parameters with back-propagation. For each training step, the teacher branch is updated in moving average approach according to the parameters of student branch. Only the feature maps predicted by the teacher decoder are given as the input of the \ac{F2D} mechanism while the depth branch affiliates optical flow prediction by feeding student decoder feature maps via \ac{D2F} mechanism. After being trained, the model discards the student decoder and the \ac{F2D} mechanism extracts optical flow maps from teacher decoder.

\subsection{Dual-head Mechanism}
\label{sec:dual-head}

Self-attention based framework \cite{2018BERT,attentionisallyouneed} avoids the locality in recurrent operations by making full use of global dependencies between input and output. The self-attention framework with multi-head attention mechanism can capture information from different representation subspaces and further achieve better performance. 
Zheng et al. \cite{zheng2022residual} proposed a \ac{RA} module to decouple different motions of nearby objects with a multi-head mask. While in this paper, inspired by the use of multi-head attention and following the idea of \textbf{divide-and-conquer}, we adopted dual-head disentanglement for \ac{D2F} mechanism to further improve the optical flow estimation performance by decoupling the rigid and non-rigid motions.

The dual-head mechanism is applied to divide optical flow into rigid and non-rigid motion field representation. As shown in Fig.~\ref{fig:mechanism}, each dual-head mask $\bm{M}^{i}$ is updated from the combination of the encoder feature map $\bm{\Phi}_{en}^{i}$ and the upsampled previous dual-head mask $\bm{M}^{i-1}$. While the first mask $\bm{M}^{1}$ is updated from the encoder feature map $\bm{\Phi}_{en}^{1}$ only:
\begin{equation}
    \bm{M}^{i}
    = \left\{ 
    \begin{array}{cl}
    \varsigma \left(\mathrm{\mathcal{C}^{3}}
    \left(
    \mathrm{ReLU}\left(\mathrm{\mathcal{C}^{3}} \left(\bm{\Phi}_{\rm en}^{i}\right)\right)
    \oplus
    \mathrm{up}\left(\bm{M}^{i}\right)
    \right)\right)
    , &i = 2,3,4
    \\
    \varsigma \left(\mathrm{\mathcal{C}^{3}}
    \left(
    \mathrm{ReLU}\left(\mathrm{\mathcal{C}^{3}} \left(\bm{\Phi}_{\rm en}^{i}\right)\right)
    \right)\right), &i = 1
    \end{array}
    \right.
    \label{dual-head}
\end{equation}
where $\rm \varsigma (\cdot)$ and $\rm ReLU(\cdot)$ respectively denotes sigmoid and ReLU operation.


The dual-head feature map $\bm{\Psi}_{t}^{i}$ of image $\bm{I}_{t}$ is generated by the tensor product of $\bm{M}_{t}^{i}$ and $\bm{\Phi}_{t}^{i}$. Meanwhile, $\bm{M}_{t+1}^{i}$ and $\bm{\Phi}_{t+1}^{i}$ are first warped by $\bm{F}_{\rm up}^{i-1}$ into $\overline{\bm{M}}_{t}^{i}$ and $\overline{\bm{\Psi}}_{t}^{i}$ before conducting tensor product operation.

\subsection{Experiments Implementation}
\label{exp implement}

In this section, we validate the improvement of (1) the cross-task collaboration, (2) the dual-head mechanism, and (3) the \ac{EMA} approach to depth evaluation and optical flow prediction. The network is trained on KITTI's data spit of Eigen et al. \cite{eigen2015predicting} following Zhou et al.'s \cite{zhou2017unsupervised} pre-processing. We train the network in the following steps:

(1) The feature encoder and the depth branch are trained with the learning rate of $10^{-4}$ for 20 epochs.

(2) The optical flow decoder branch and \ac{D2F} blocks are further connected into the network. The network is trained with the learning rate of $10^{-4}$ for another 20 epochs.

(3) Finally, we add the \ac{F2D} block for depth evaluation assistance. The network is trained with the learning rate of $10^{-5}$ for 5 epochs.

We trained our network using depth loss $L_{\rm depth}$ and optical flow loss $L_{\rm flow}$ following \cite{godard2019digging} and \cite{sun2018pwc}, respectively. Adam optimizer \cite{kingma2014adam} is used in all of the training steps. The size of the images is $1024 \times 320$. We validate the depth estimation result on the Eigen
et al.’s testing split, and the optical flow prediction result on KITTI 2015 training set.

\section{RESULTS}

\subsection{Ablation Study}

\begin{table*}[htpb]
    \centering
    \begin{tabular}{c|c|c|c|ccccccc|ccc}
        \hline
        \multirow{2}{*}{Model} & \multirow{2}{*}{Components} & \multirow{2}{*}{Parameters} & \multirow{2}{*}{FPS} &  \multicolumn{7}{|c|}{Depth} & \multicolumn{3}{c}{Optical Flow} \\
        \cline{5-14}
         & & & & abs rel$\downarrow$ & sq rel$\downarrow$ & rms$\downarrow$ & log rms$\downarrow$ & $\rm \delta_{1}\uparrow$ & $\rm \delta_{2}\uparrow$ & $\rm \delta_{3}\uparrow$ & epe$\downarrow$ & epe noc$\downarrow$ & F1(\%)$\downarrow$\\
        \hline
        \uppercase\expandafter{\romannumeral1} & Baseline & 19.97M & - & 0.115 & 0.885 & \textbf{4.705} & \textbf{0.190} & 0.879 & \textbf{0.961} & \textbf{0.982} & 8.20 & 4.53 & 24.35 \\
        \hline
        \uppercase\expandafter{\romannumeral2} & \uppercase\expandafter{\romannumeral1} + D2F & 18.22M & 12.8 & 0.117 & 0.954 & 4.878 & 0.195 & 0.878 & 0.960 & 0.980 & 6.73 & 3.23 & 20.65 \\
        \hline
        \uppercase\expandafter{\romannumeral3} & \uppercase\expandafter{\romannumeral2} + \ac{F2D} & 19.00M & 11.9 & 0.113 & 0.886 & 4.795 & 0.193 & 0.880 & 0.960 & 0.981 & 7.23 & 3.42 & 21.16 \\
        \hline
        \uppercase\expandafter{\romannumeral4} & \uppercase\expandafter{\romannumeral3} + \ac{EMA} & 19.00M & 11.9 & \textbf{0.112} & 0.893 & 4.788 & 0.192 & \textbf{0.882} & 0.960 & 0.981 & 7.11 & 3.30 & 21.16 \\
        \hline
        \uppercase\expandafter{\romannumeral5} & \uppercase\expandafter{\romannumeral2} + D-H & 18.78M & 8.1 & 0.117 & 1.000 & 4.859 & 0.194 & 0.880 & 0.960 & 0.980 & \textbf{6.23} & \textbf{2.88} & \textbf{19.88} \\
        \hline
        \uppercase\expandafter{\romannumeral6} & \uppercase\expandafter{\romannumeral4} + D-H & 19.55M & 7.7 & \textbf{0.112} & \textbf{0.859} & 4.798 & 0.193 & 0.879 & 0.960 & 0.981 & 6.84 & 3.04 & 20.89 \\
        \hline
    \end{tabular}
    \caption{Ablation study indicating how the mechanisms we proposed affect the parameter amount, \ac{FPS}, and prediction result of depth and optical flow. Baseline: monodepth2 \cite{godard2019digging} and PWC-Net \cite{sun2018pwc} working separately. The parameter mount is the sum of these two networks. \ac{D2F}: depth-to-flow  mechanism. \ac{F2D}: flow-to-depth mechanism. \ac{EMA}: \ac{EMA} approach. D-H: dual-head mechanism.}
    \label{abl_study}
\end{table*}

\begin{figure}[htpb]
    \centering
    \includegraphics[width=1\linewidth]{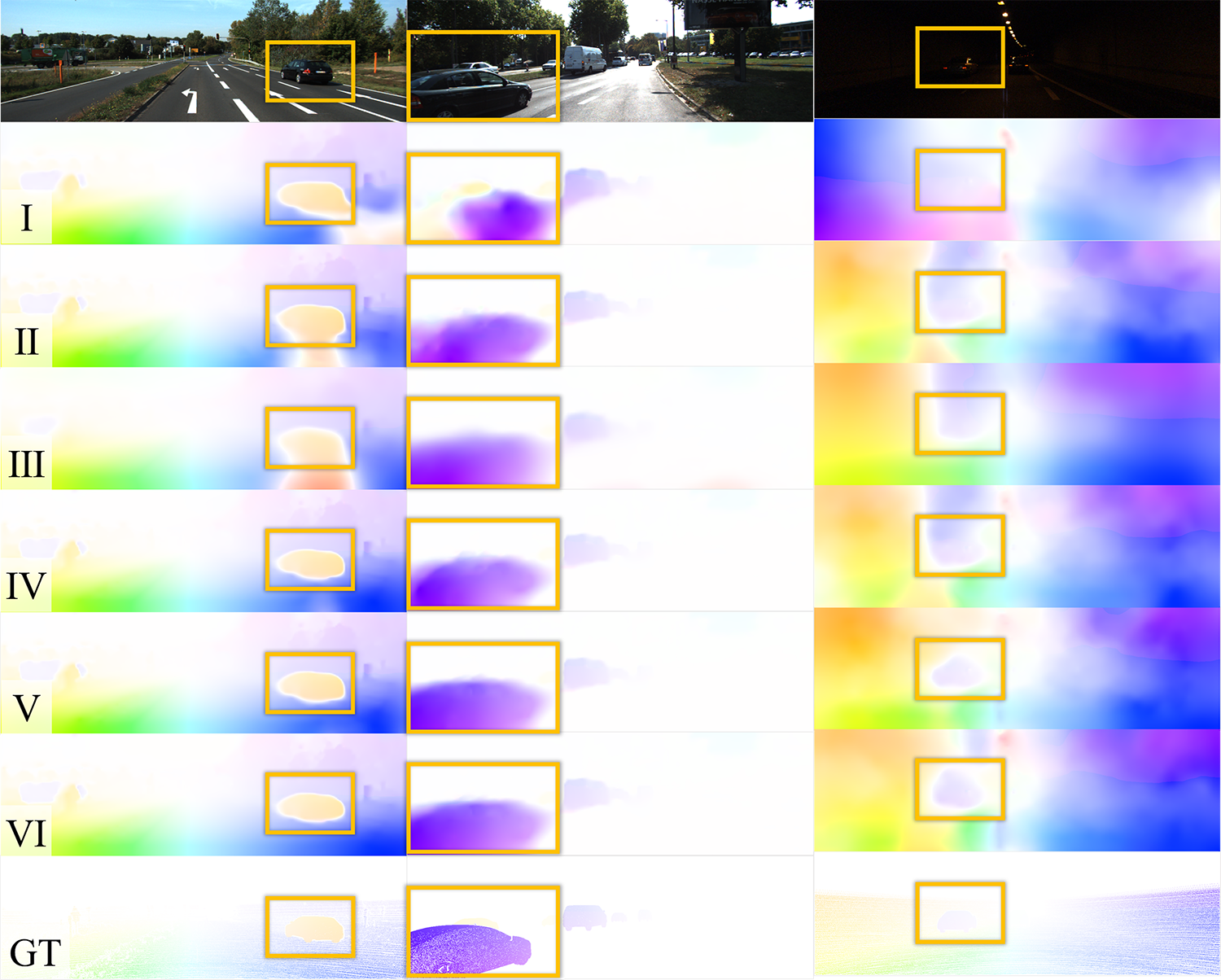}
    \caption{Illustration of the ablation study of optical flow prediction. From top to bottom: input color image and the prediction result of model \uppercase\expandafter{\romannumeral1} to \uppercase\expandafter{\romannumeral6}, and the optical flow \ac{GT}. In the orange boxes are vehicles or their respective optical flow.}
    \label{flow_abl}
\end{figure}

\begin{figure}[htpb]
    \centering
    \includegraphics[width=1\linewidth]{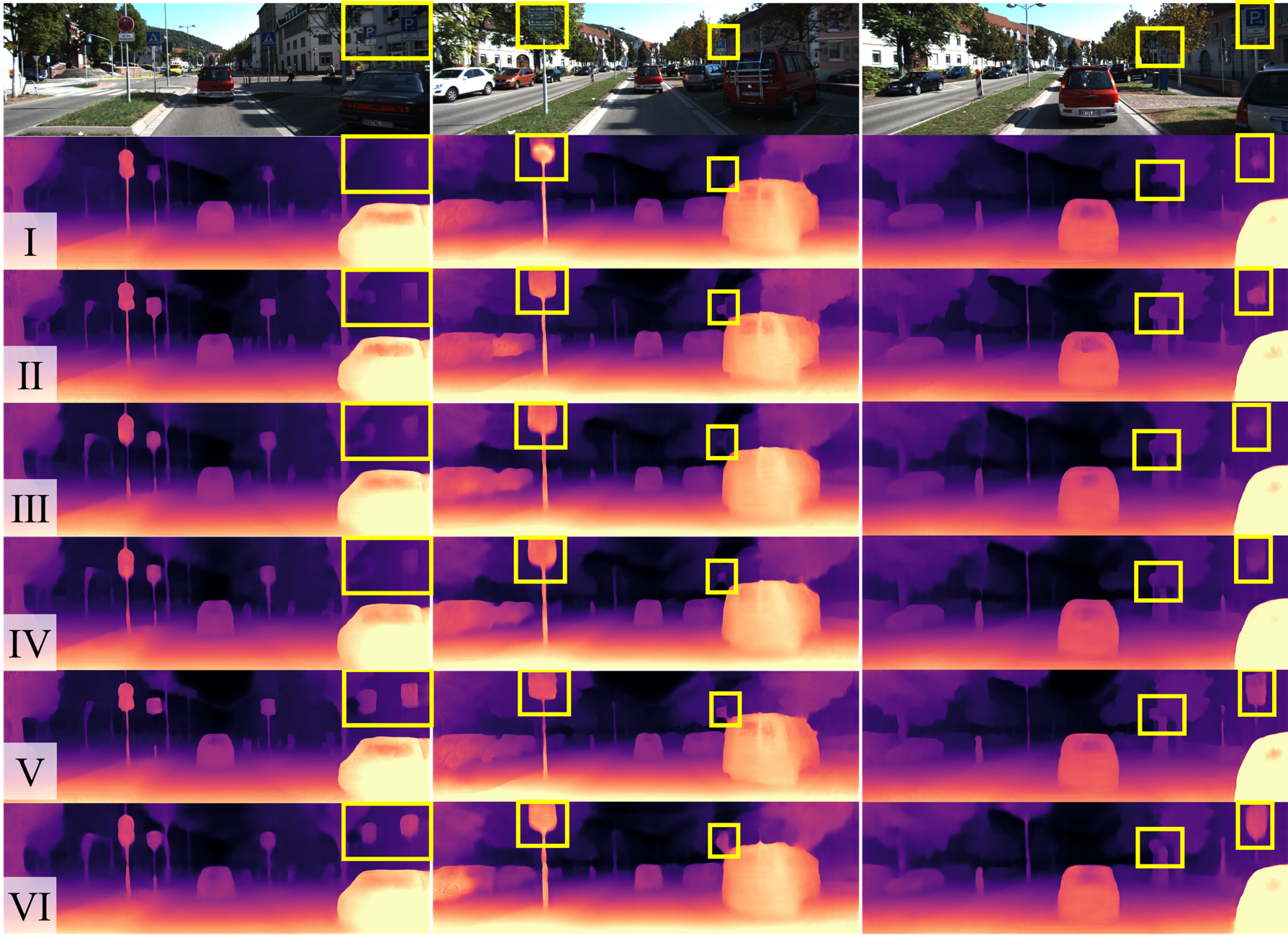}
    \caption{Illustration of the depth estimation. From top to bottom: input color image and the prediction result of model \uppercase\expandafter{\romannumeral1} to \uppercase\expandafter{\romannumeral6}. In the yellow boxes, we show some road signs that are easy to be confused with the background buildings.}
    \label{depth_abl}
\end{figure}

\begin{figure}[htpb]
    \centering
    \includegraphics[width=1\linewidth]{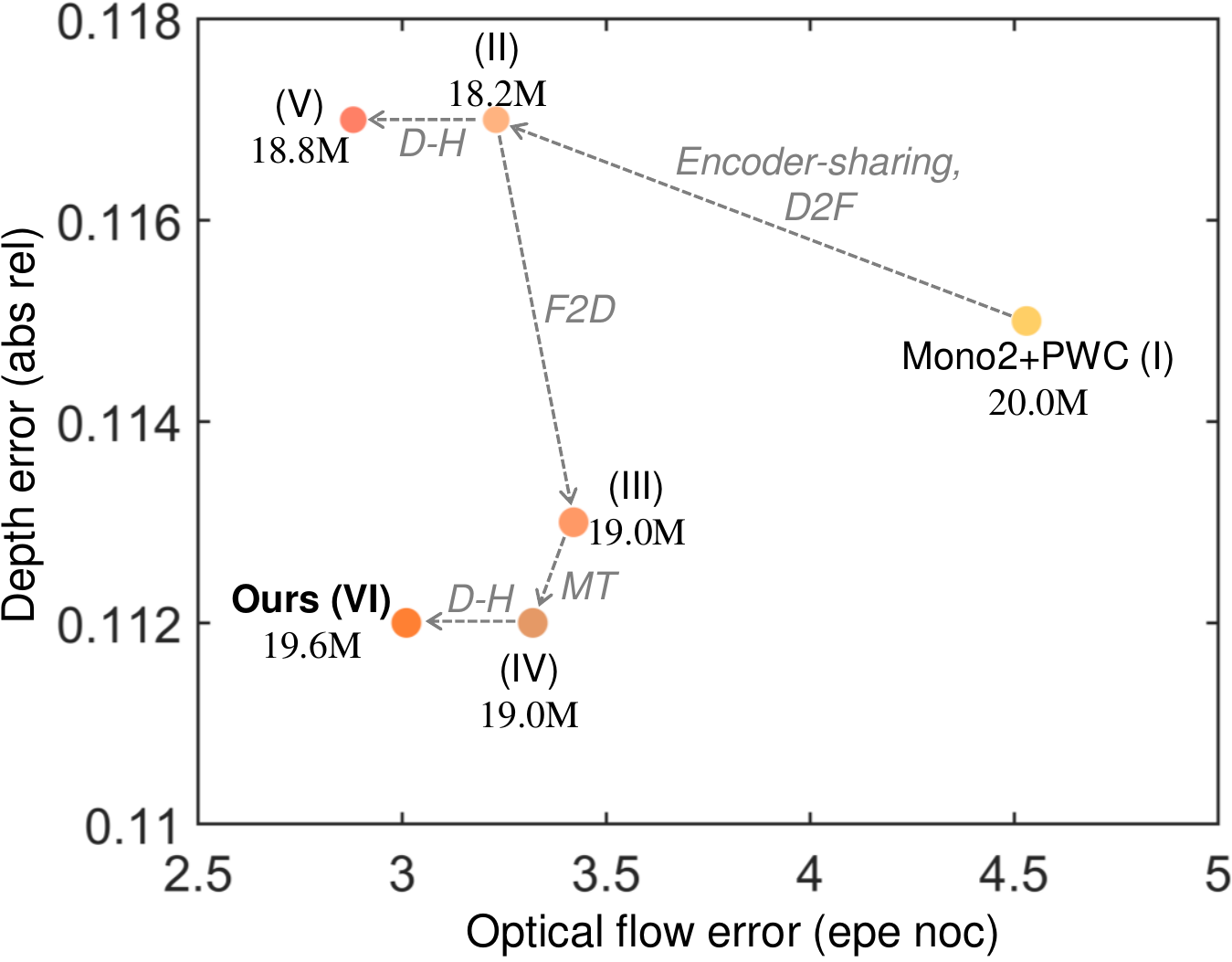}
    \caption{Bubble chart indicating the trade-off of depth and optical flow prediction accuracy, represented with absolute relative error and non-occluded pixels endpoint error, of model \uppercase\expandafter{\romannumeral1} to \uppercase\expandafter{\romannumeral6}. The parameter mounts of models are shown with the size of the bubbles.}
    \label{bubble_ablation}
\end{figure}

To validate the effect of each component of depth and optical flow prediction network, we tried different combinations of \ac{D2F} mechanism, \ac{F2D} mechanism, \ac{EMA} approach, and dual-head mechanism in experiments. Tab.~\ref{abl_study} shows the prediction results of the experiments with the various combinations, marked as \uppercase\expandafter{\romannumeral1} to \uppercase\expandafter{\romannumeral6}.  Fig.~\ref{flow_abl} and \ref{depth_abl} illustrate the effect of \ac{D2F} and dual-head mechanism to the result of optical flow and depth estimation. Fig.~\ref{bubble_ablation} further illustrates the trade-off between depth and optical flow prediction and adding different components to the network architecture. 

\ac{D2F} mechanism can significantly improve the accuracy of optical flow prediction, which can be proved by comparing experiment \uppercase\expandafter{\romannumeral1} and \uppercase\expandafter{\romannumeral2} in Tab.~\ref{abl_study} and Fig.~\ref{flow_abl}. Furthermore, it can be seen from Fig.~\ref{depth_abl} that \ac{D2F} mechanism can also support the depth branch telling the road signs from the buildings. This might be because objects like road signs are closer to the camera and can generate optical flow different from the distant buildings, which can be detected by the optical flow branch.

The information generated from optical flow prediction facilitates depth estimation via \ac{F2D} mechanism. This could be seen from the comparison of experiment \uppercase\expandafter{\romannumeral1}, \uppercase\expandafter{\romannumeral2}, and \uppercase\expandafter{\romannumeral3} in Tab.~\ref{abl_study}. Experiment \uppercase\expandafter{\romannumeral3} and \uppercase\expandafter{\romannumeral4} also demonstrates that \ac{EMA} approach can benefit both depth and optical flow prediction. 

Experiment \uppercase\expandafter{\romannumeral5} and \uppercase\expandafter{\romannumeral6}, respectively compared to experiment \uppercase\expandafter{\romannumeral2} and \uppercase\expandafter{\romannumeral4}, prove that dual-head mechanism supports the optical flow prediction accuracy. The supporting role of dual-head mechanism can also be seen in Fig.~ \ref{flow_abl} and \ref{depth_abl}. As shown in Fig.~\ref{flow_abl}, comparing to baseline, with or without \ac{D2F} mechanism's assistance, adding dual-head mechanism to the architecture can significantly improve the network's ability to recognize moving objects like vehicles -- even in the dark tunnel. Fig.~\ref{depth_abl} shows that during depth estimation, the network with dual-head mechanism can also help distinguish between buildings and road signs.

Fig.~\ref{bubble_ablation} tells how the different components affect the two tasks of the network. \ac{D2F} mechanism significantly improves optical flow prediction while causing the depth estimation accuracy slightly drops. A similar effect was achieved with \ac{F2D}, enhances the accuracy of depth prediction while mildly reducing the effect of optical flow prediction. \ac{EMA} approach achieves a positive effect on both tasks and dual-head mechanism 
facilitates optical flow prediction with little impact on depth estimation.

\subsection{Depth and Optical Flow Evaluation}

\begin{table*}[h!]
    \centering
    \begin{tabular}{c|c|c|c|ccccccc|ccc}
        \hline
        Method & Parameters & FPS & Scheme & abs rel$\downarrow$ & sq rel$\downarrow$ & rms$\downarrow$ & log rms$\downarrow$ & $\rm \delta_{1}\uparrow$ & $\rm \delta_{2}\uparrow$ & $\rm \delta_{3}\uparrow$ & epe$\downarrow$ & epe noc$\downarrow$ & F1(\%)$\downarrow$ \\
        
        \hline
        
        zhou et al. \cite{zhou2017unsupervised} & - & - & - & 0.183 & 1.595 & 6.709 & 0.270 & 0.734 & 0.902 & 0.959 & - & - & - \\
        
        monodepth2 \cite{godard2019digging} & - & - & - & \underline{0.115} & \underline{0.882} & \underline{4.701} & \textbf{0.190} & \underline{0.879} & \textbf{0.961} & \textbf{0.982} & - & - & - \\
        
        \hline
        
        Back2Future & - & - & - & - & - & - & - & - & - & - & 7.04 & - & 24.21 \\
        Unflow \cite{meister2018unflow} & - & - & - & - & - & - & - & - & - & - & 8.10 & - & 23.27 \\
        
        \hline
        
        DF-Net \cite{zou2018df} & 151.15M & - & (a) & 0.150 & 1.124 & 5.507 & 0.223 & 0.806 & 0.933 & 0.973 & 8.98 & - & 26.01 \\
        
        GL-Net \cite{chen2019self} & - & - & (a) &  0.135 & 1.070 & 5.230 & 0.210 & 0.841 & 0.948 &  0.980 & 8.35 & 4.86 & - \\
        
        CC \cite{ranjan2019competitive} & 59.77M & 7.8 & (a) &  0.140 & 1.070 & 5.326 & 0.217 & 0.826 & 0.941 & 0.975 & \underline{6.27} & 4.17 & 29.15 \\
        
        DOP \cite{wang2020unsupervised} & 59.77M & 7.9 & (a) &  0.140 & 1.068 & 5.255 & 0.217 & 0.827 & 0.943 & 0.977 & 6.66 & - & 23.04 \\



        GeoNet \cite{yin2018geonet} & 118.52M & 0.2 & (b) &  0.147 & 0.936 & \textbf{4.348} & 0.218 & 0.810 & 0.941 & 0.977 & 10.81 & 8.05 & - \\

        Yang. \cite{yang2021unsupervised} & - & - &  (d) &  0.139 & 1.297 & 5.879 & 0.223 & 0.827 & 0.936 & 0.979 & 9.87 & 6.45 & - \\
        \hline
        
        Ours (\uppercase\expandafter{\romannumeral5}) & 18.78M & 8.1 & (e) & 0.117 & 1.000 & 4.859 & 0.194 & \textbf{0.880} & \underline{0.960} & 0.980 & \textbf{6.23} & \textbf{2.88} & \textbf{19.88} \\
        
        Ours (\uppercase\expandafter{\romannumeral6}) & 19.55M & 7.7 & (e) & \textbf{0.112} & \textbf{0.859} & 4.798 & \textbf{0.193} & \underline{0.879} & \underline{0.960} & \underline{0.981} & 6.83 & \underline{3.01} & \underline{20.88} \\

        \hline
    \end{tabular}
    \caption{Depth estimation comparison. Top row: unsupervised monocular depth estimation methods. Middle top row: unsupervised optical flow prediction methods. Middle bottom row: multi-task estimation methods. Bottom row: our method.}
    \label{depth_comparison}
\end{table*}

\begin{figure*}[htpb]
    \centering
    \includegraphics[width=0.92\linewidth]{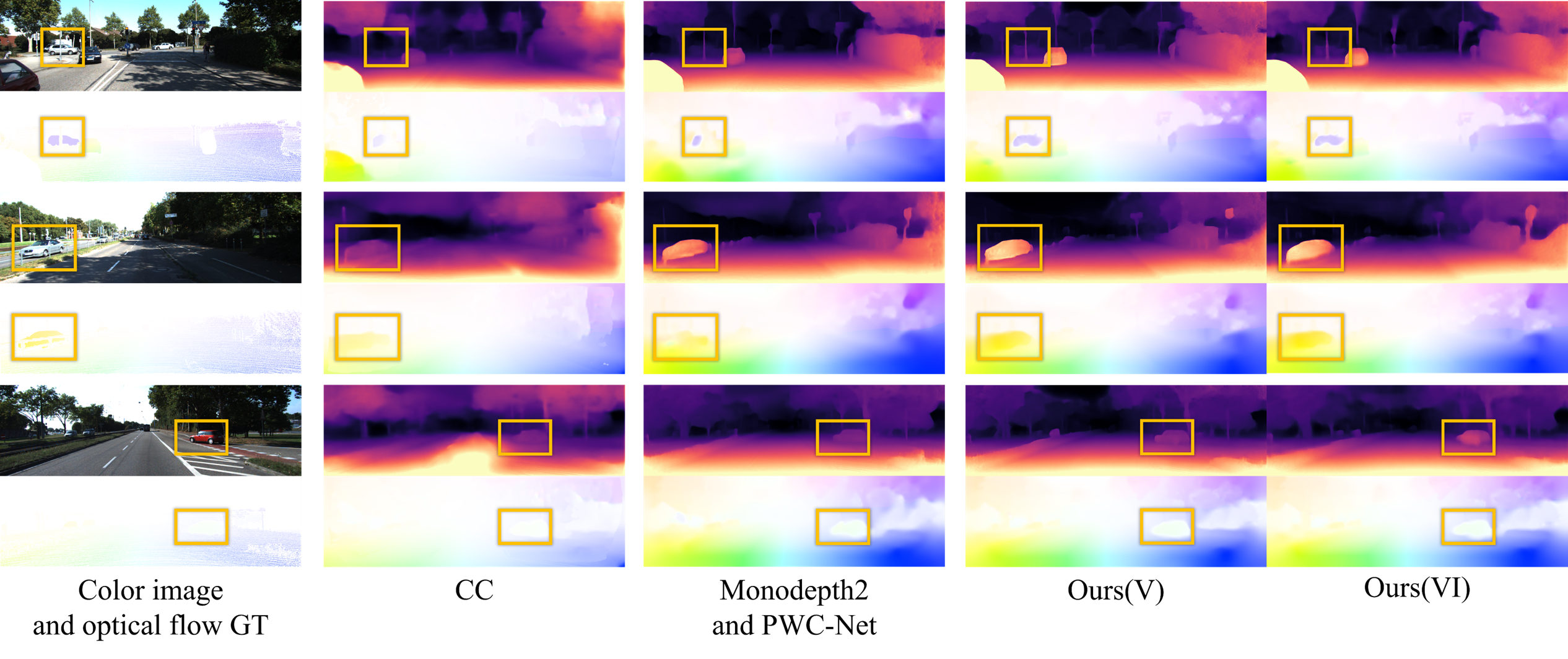}
    \caption{Depth and optical flow prediction illustration comparison between CC \cite{ranjan2019competitive}, monodepth2 \cite{godard2019digging}+PWC-Net \cite{sun2018pwc}, and our method (model \uppercase\expandafter{\romannumeral5} and \uppercase\expandafter{\romannumeral6}). Column (a) shows the target image and its optical flow \ac{GT}. Column (b), (c), and (d) shows the depth and optical flow prediction results of CC, monodepth2 and PWC-Net, and our method.}
    \label{compare}
\end{figure*}

\begin{figure}[htpb]
    \centering
    \includegraphics[width=0.95\linewidth]{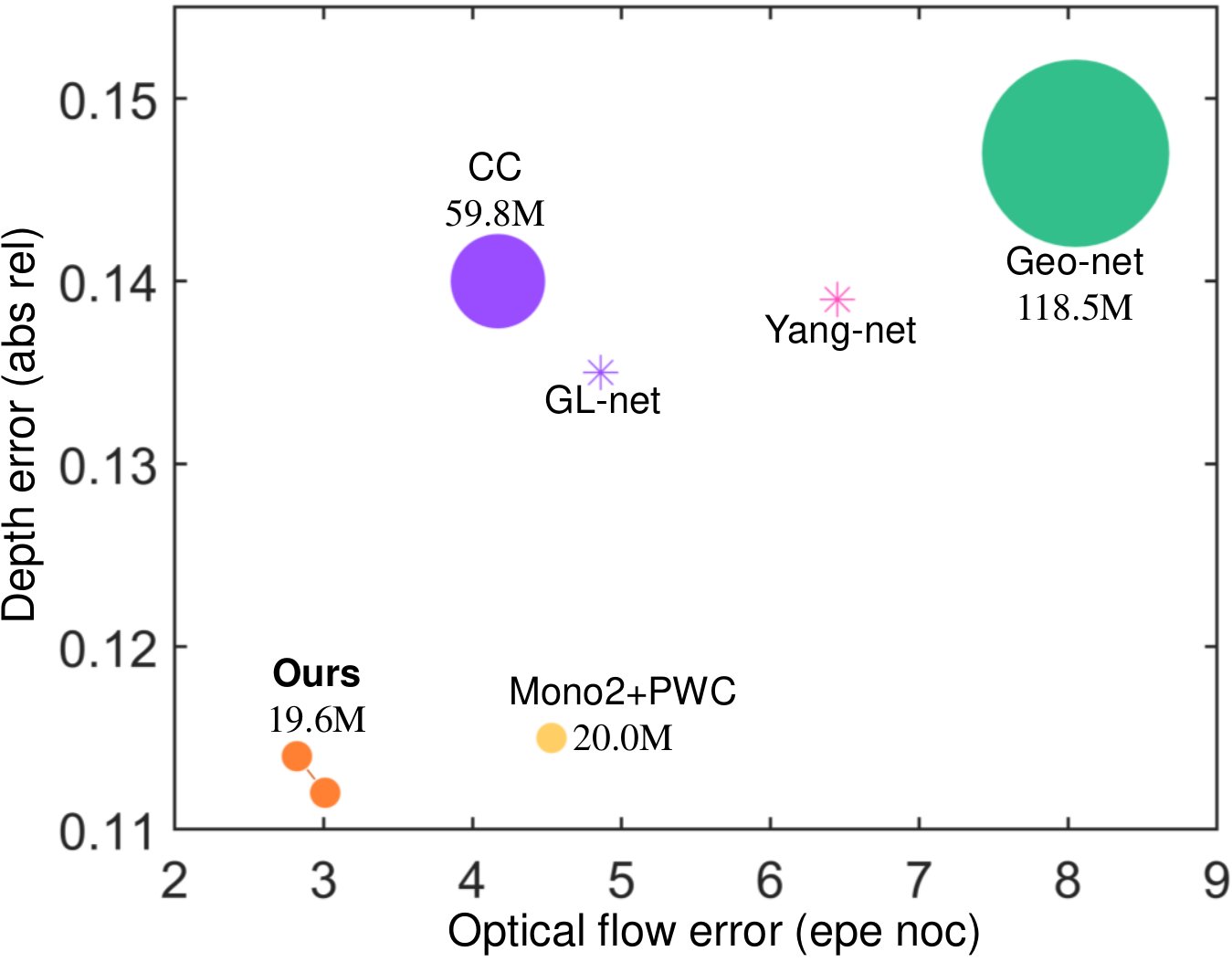}
    \caption{Bubble chart comparing the trade-off of depth estimation, optical flow prediction, and parameter mount between different methods. The horizontal and vertical axis represents optical flow prediction non-occluded pixels endpoint error and depth prediction absolute relative error, while the size of the bubbles shows the parameter mounts.}
    \label{bubble_compare}
\end{figure}

The comparisons of monocular depth and optical flow estimation between our approach and other methods are summarized in Tab.~\ref{depth_comparison}, Fig.~\ref{compare}, and Fig.~\ref{bubble_compare}. 

In Tab.~\ref{depth_comparison} and Fig.~\ref{bubble_compare}, we take our models in experiment \uppercase\expandafter{\romannumeral5} and \uppercase\expandafter{\romannumeral6} of Tab.~\ref{abl_study} to compare with other baselines. Our models achieve better or comparable results for depth and optical flow estimation. Moreover, benefit from shared encoder and small components, our models contain small amount of parameters and thus achieve higher or comparable \ac{FPS}. Moreover, as shown in Fig.~\ref{compare}, in both depth and optical flow prediction, our method achieves significant improvement of detecting vehicles.







\section{CONCLUSION}

In this paper, we design a multi-task network that achieves mutual assistance between the two tasks of optical flow and depth prediction through bi-directional information interaction. Specifically, we design D2F and F2D modules for cross-domain information interaction and apply the \ac{EMA} method to the training of the network. We further enhance the network's ability to distinguish objects during multi-task prediction by the applying dual-head mechanism to the optical flow branch. Besides improving the mean accuracy of depth and optical flow prediction, our method significantly facilitates the ability to tell objects like road signs and vehicles from the scene. In the future, our work can be applied to the visual perception of autonomous vehicles and the navigation of surgical robots.

\bibliographystyle{IEEEtran}
\bibliography{RAL.bib}

\end{document}